\documentclass[a4paper, times, 10pt,twocolumn]{article}
\usepackage[top=4.9cm,bottom=3.7cm,left=1.5cm,right=1.5cm]{geometry}
\usepackage{ICMLC}
\usepackage{times}
\usepackage{graphicx}
\usepackage{indentfirst}
\usepackage{latexsym}
\usepackage{amsfonts}
\usepackage{amsmath}
\usepackage{multirow}
\usepackage{comment}

\usepackage[margin=8pt,font=footnotesize,labelfont=bf,labelsep=period
]{caption}

\pdfpagewidth=\paperwidth
\pdfpageheight=\paperheight
\begin{document}

\title{An Access Control Method with Secret Key for Semantic Segmentation Models}  

\author{\bf{\normalsize{Teru Nagamori${^1}$, Ryota Iijima${^1}$, Hitoshi Kiya${^1}$}}\\ 
\\
\normalsize{$^1$ Department of Computer Science, Tokyo Metropolitan University, Hino, Tokyo, Japan}\\
\normalsize{E-MAIL: nagamori-teru@ed.tmu.ac.jp, iijima-ryota@ed.tmu.ac.jp, kiya@tmu.ac.jp}\\
\\}

\maketitle \thispagestyle{empty}

\begin{abstract}
   {A novel method for access control with a secret key is proposed to protect models from unauthorized access in this paper. We focus on semantic segmentation models with the vision transformer (ViT), called segmentation transformer (SETR). Most existing access control methods focus on image classification tasks, or they are limited to CNNs. By using a patch embedding structure that ViT has, trained models and test images can be efficiently encrypted with a secret key, and then semantic segmentation tasks are carried out in the encrypted domain. In an experiment, the method is confirmed to provide the same accuracy as that of using plain images without any encryption to authorized users with a correct key and also to provide an extremely degraded accuracy to unauthorized users.}
\end{abstract}

\section{Introduction}
Deep neural networks (DNNs) have been used widely in various applications such as semantic segmentation, image classification, and object detection \cite{krizhevsky2017imagenet,Review, Object_Detection}. Training models is a difficult task in general, because it needs a very large amount of data, computational resources, and excellent algorithms. Considering the expertise, and cost required for training models, they should be protected as a kind of intellectual property.\par
Two approaches: ownership verification and access control, have been considered to protect intellectual property of models. The former aims to identify the ownership of the models, and the latter aims to restrict the use of the models from unauthorized access \cite{kiya2022overview}. The ownership verification methods have been inspired by watermarking, where the watermark embedded in models is used to verify the ownership of the models \cite{uchida2017embedding,Visual_Decoding,Turning,maung2021piracy}. However, the embedded watermark does not have the ability to restrict the execution of the models. Thus, unauthorized users can utilize the models for their own benefit, or use them in adversarial attacks \cite{Explaining}. Therefore, in this paper, we discuss access control to prevent models from unauthorized access. \par 
By encrypting images or feature maps, unauthorized users without a correct secret key cannot use a model \cite{maungmaung_kiya_2021,ito2021access,teru,Maung_IEEE, Ito_apsipa}. However, the conventional methods with a secret key are required to train a model with a key. Therefore, the trained model has to be retrained with images encrypted with a different key if we want to update a key. In addition, the conventional methods cannot provide the same accuracy as that of using plain images without any encryption to authorized users with a correct key. Moreover, any access control methods have never been considered for semantic segmentation models with the vision transformer (ViT) \cite{ViT}. \par
Accordingly, we propose a novel access control method for semantic segmentation models with the vision transformer, called segmentation transformer (SETR) \cite{SETR} for the first time. In experiments, the proposed method is verified to allow authorized users not only to obtain the same accuracy as that of using plain images, but to also update a secret key easily.
\section{Related Work}
\subsection{Access Control}
Access control aims to protect functionalities of models from unauthorized access. Therefore, protected models are required not only to provide a high accuracy to authorized users, but to also provide a low accuracy to unauthorized users (see Fig. \ref{access_control}). In addition, the protected models have to be robust against various attacks.
Most access control 
methods focus on protecting models in image classification tasks \cite{maungmaung_kiya_2021,Iijima}, but they cannot be applied to semantic segmentation tasks, because results with a pixel-level resolution are required in semantic segmentation tasks. An existing access control method in semantic segmentation tasks is carried out by encrypting feature maps of CNNs \cite{ito2021access}. However, since this method is limited to CNNs with feature maps, it cannot be applied to models with ViT, which have a higher performance than that of CNN-based models. \par
Accordingly, we propose an access control method for models with ViT for the first time in this paper.

\subsection{Segmentation Transformer}
\begin{figure*}[tb]
    \centering
    \includegraphics[bb=0 0 952 188,scale=0.5]{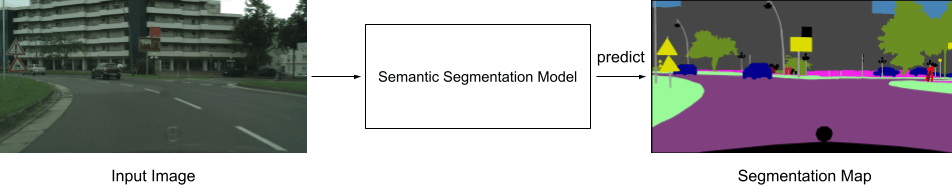}
    \caption{Overview of Semantic Segmentation}
    \label{semantic_segmentaion}
\end{figure*}

The purpose of semantic segmentation is to classify objects in a pixel-level resolution. Figure \ref{semantic_segmentaion} illustrates an overview of semantic segmentation. A model predicts a segmentation map from an input image, where each pixel in the segmentation map represents a class label.
The segmentation transformer (SETR) is the first transformer-based model proposed for semantic segmentation \cite{SETR}. This model has been inspired by ViT, which has a high performance in image classification tasks. \par
\begin{figure}[tb]
    \centering
    \includegraphics[bb=0 0 776 746, scale=0.3]{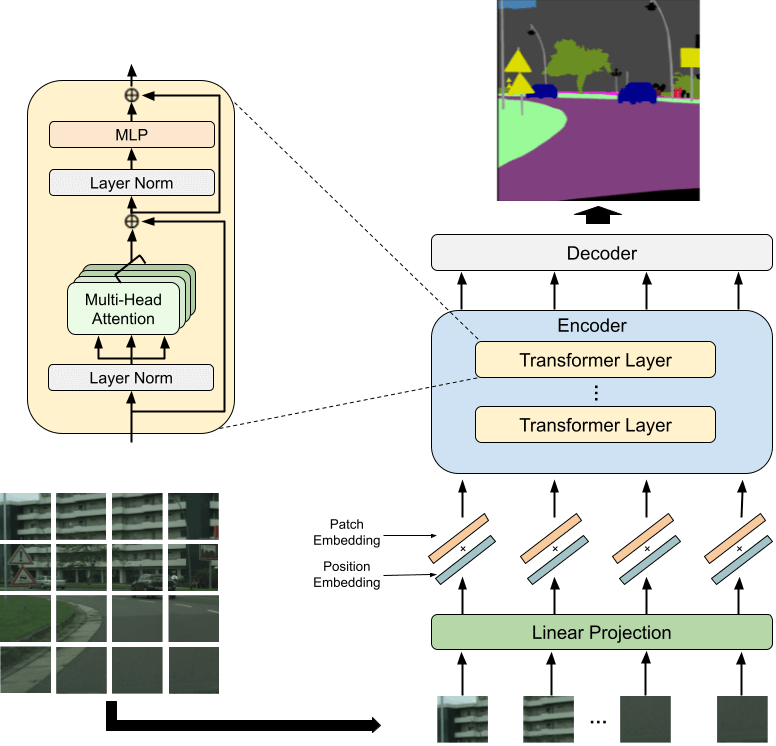}
    \caption{Architecture of Segmentation Transformer \cite{SETR}}
    \label{SETR}
\end{figure}

Figure \ref{SETR} shows the architecture of SETR. Encoder is the same as ViT. Since the encoder in ViT receives only 1-D vectors as an input, an image is divided into patches. Each patch is then flattened and converted to a 1-D vector. Two embeddings are used to understand the location information of separated patches in an original image, and each patch is mapped to learnable dimensions. The former, called position embedding, embeds location information about where each patch is in an original image. The latter, called patch embedding, maps each patch to learnable dimensions using a matrix. By using these two types of embedding, learning is possible even when an image is divided into patches while keeping location information. After that, the feature representation obtained by the encoder in ViT is inputted to a decoder, which outputs a segmentation map of the same size as an input image. \par


\section{Proposed Method}
\subsection{Overview}
\begin{figure}[tb]
    \centering
    \includegraphics[bb=0 0 816 823,scale=0.3]{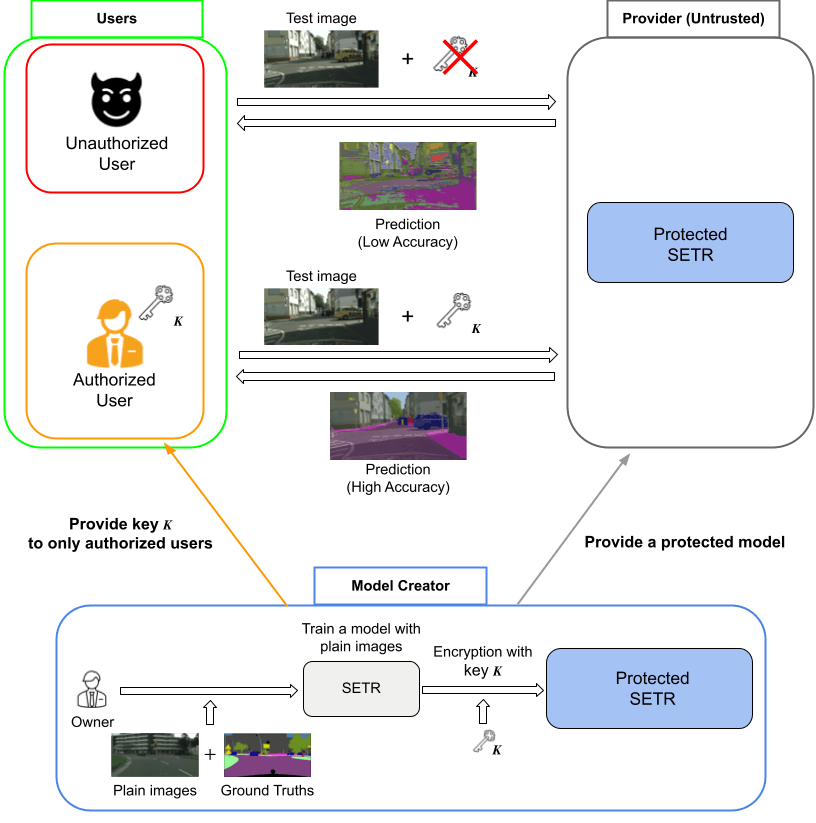}
    \caption{Overview of Access Control}
    \label{access_control}
\end{figure}

Figure \ref{access_control} shows an overview of the proposed access control method for protecting trained models from unauthorized access. The method is summarized as bellow. \par
First, a model creator trains a model with plain images. Next, after training a model, the model is encrypted with a secret key. After then, the secret key is provided to authorized users, and the protected model is provided to a provider, who may not be trusted. Authorized users can obtain a high performance from the model. In contrast, unauthorized users without a key cannot obtain a high performance.

\subsection{Encryption Method}
\label {proposed_method}
ViT utilizes patch embedding and position embedding. In this paper, we use two unique properties due to these embeddings. In SETR, an image $x$ with a dimension of $h \times w \times c$ is divided into $N$ patches with a patch size of $p \times p$ where $h, w$ and $c$ are the height, width, and the number of channels of $x$. Namely, $N$ is given as $hw/p^2$. After that, each patch is flattened as $x_{p}^{i}$, and it is linearly mapped to a vector with dimensions of $D$ and a learnable matrix \textbf{E} as
\begin{align}
    z_{0}^i =& x_{p}^{i}\mathbf{E} + \mathbf{E_{pos}^i},\notag\\
    i \in& \left\{1,2,...,N\right\}, \ x_{p}^{i} \in \mathbb{R}^{p^{2}c},\\
     \mathbf{E} \in& \mathbb{R}^{(p^{2}c) \times D}, \ 
     \mathbf{E_{pos}^i} \in \mathbb{R}^{D}.\notag
\end{align}
where $x_{p}^{i}$ is the $i$-th patch of a train image, $\mathbf{E_{pos}^i}$ is the position information of $x_{p}^{i}$, and $z_{0}^i$ is the $i$-th embedded patch.\par
The proposed method is carried out in accordance with the above relation.

\subsubsection{Model Encryption}
In the proposed method, \textbf{E} is transformed with a secret key $K$ after training a model as follows.
\begin{itemize}
    \item[1)] Generate a matrix $\mathbf{E_{enc}}$ with a secret key $K$ as
    $$
    \mathbf{E_{enc}} =
    \begin{bmatrix} 
          k_{(1,1)} & k_{(1,2)} & \dots & k_{(1,p^{2}c)} \\
          k_{(2,1)} & k_{(2,2)} &\dots  & k_{(2,p^{2}c)} \\
          \vdots & \vdots & \ddots & \vdots \\
          k_{(p^{2}c,1)} & k_{(p^{2}c,2)} & \dots & k_{(p^{2}c,p^{2}c)}\\
    \end{bmatrix},
    $$
    \begin{align*}
    k_{(i,j)} &\in \mathbb{R},\ i,j \in \left\{1, \dots, p^{2}c\right\}, \\
    \mathbf{E_{enc}} &\in \mathbb{R}^{(p^{2}c) \times (p^{2}c)}, \rm{and} \ \mathrm{det}\mathbf{E_{enc}} \neq 0.
    \end{align*}
    \item[2)]Multiply $\mathbf{E_{enc}}$ and \textbf{E} to obtain $\mathbf{E'}$ as
    \begin{equation}
            \mathbf{E'} = \mathbf{E_{enc}}\mathbf{E}, \ \mathbf{E'} \in \mathbb{R}^{(p^{2}c) \times D}.
    \end{equation}
    \item[3)]Replace \textbf{E} in Eq.(1) with $\mathbf{E'}$ as a new patch embedding to encrypt a model.
\end{itemize}

\subsubsection{Test Image Encryption}
We assume that an authorized user has key $K$ and the user can generate $\mathbf{E_{enc}}$ with $K$ as well. Accordingly, a new matrix $\mathbf{\hat{E}_{enc}}$ can be computed by
\begin{equation}
    \mathbf{\hat{E}_{enc}} =\mathbf{E_{enc}^{-1}}.
\end{equation}
An encrypted test image $\hat{x}$ is produced by an authorized user as follows.
\begin{itemize}
    \item[(a)] Divide an input image $x$ into blocks with a patch size of $p \times p$ such that $\left\{B_{1},\dots ,B_{N}\right\}$.
    \item[(b)]Flatten each block $B_{i}$ into a vector $b_{i}$ such that
     \begin{equation}
        b_{i} = [b_{i}(1), \dots , b_{i}(p^{2}c)].
    \end{equation}
    \item[(c)]Generate an encrypted vector $\hat{b_{i}}$ by multiplying $b_{i}$ by $\mathbf{\hat{E}_{enc}}$ as
    \begin{equation}
                \hat{b_{i}} = b_{i}\mathbf{\hat{E}_{enc}}, \ \hat{b_{i}} \in \mathbb{R}^{p^{2}c}.
    \end{equation}
    
    \item[(d)]Concatenate the encrypted vectors into an encrypted image $\hat{x}$ with a dimension of $h \times w \times c$.
    \end{itemize}
As a result, when $\hat{x}$ is applied to the encrypted model, the embedded patch becomes
\begin{align*}
    z_{0}^i &= \hat{x}_{p}^i\mathbf{E'} + \mathbf{E_{pos}^i}\\
     &= x_{p}^{i}\mathbf{E} + \mathbf{E_{pos}^i},
\end{align*}
where $\hat{x}_{p}^i$ is the $i$-th patch of an encrypted test image.\par
From the above equation, $z_{0}^i$ produced with the above procedure can maintain the same value as that of no protected models under the use of models encrypted with key $K$. In contrast, the encrypted models provide a degraded performance to unauthorized users without key $K$.

\section{Experimental Result}
\subsection{Setup}
We evaluated the proposed method under the use of two datasets and the training conditions used in \cite{SETR}. \par
The first dataset was Cityscapes \cite{Cityscapes}, which is an urban scene dataset with 19 object categories. It consists of 5000 images with a resolution of 2048 $\times$ 1024 in total, and the images were divided into 2975, 500 and 1525 ones for training, validation, and testing, respectively. The another dataset was ADE20K \cite{ADE20k}, which is a benchmark for scene analysis with 150 categories. It consists of 20210, 2000 and 3352 images for training, validation and testing. In the experiment, training and validation images were used for training and testing, respectively. \par
As data augmentation techniques, we utilized random resize with a ratio between 0.5 and 2, random cropping (768 $\times$ 768 for Cityscapes and 512 $\times$ 512 for ADE20K), and random horizontal flip to training models. As a learning parameter, for Cityscapes, we set batch size to 8, total iteration to 80k, and initial learning rate to 0.01. In contrast, for ADE20K, we set batch size to 16, total iteration to 160k, and initial learning rate to 0.001. For all experiments on the two datasets, we utilized a polynomial learning rate decay schedule \cite{Zhao_2018_ECCV} and used SGD for optimization. Momentum and weight decay were set to 0.9 and 0, respectively. \par
In SETR, the encoder has two variations: T-base and T-large. T-base is a small model with 12 transformer layers and 768 hidden layers, and T-large is a large model with 24 transformer layers and 1024 hidden layers. In addition, there are three types of decoders: \textit{Na\"{i}ve}, which employs a simple two-layer network followed by a bilinear upsampler to restore an original resolution; \textit{PUP}, which alternates between convolution layers and upsampling operations; and \textit{MLA}, which uses multi-stage features like a feature pyramid network.\par
In this experiment, we used models with a T-base encoder and pre-trained with DeiT \cite{Deit} weights for Cityscapes and models with a T-large encoder and pre-trained with ViT weights for ADE20K. Besides, we also used all three types of decoders.\par
For the loss function, we not only used pixel-wise cross-entropy loss for full-resolution output of a decoder, but also added auxiliary segmentation loss to different transformer layers for each decoder: \textit{Na\"{i}ve} $= (z_{10},z_{15},z_{20})$, \textit{MLA} $= (z_{6},z_{12},z_{18},z_{24})$, \textit{PUP} $= (z_{10},z_{15},z_{20},z_{24})$, where $z_i$ represents the ${i}$-th transformer layer. \par
As an evaluation metric, we used mean Intersection over Union ($mIoU$), which is an average of Intersection over Union ($IoU$) for each class defined as
\begin{equation}
            IoU = \frac{TP}{FP + FN + TP},
\end{equation}
where $FP$, $FN$, and $TP$ are  false positive, false negative and true positive values calculated from a predicted full-resolution output and ground truth, respectively. When a IoU value is closer to 1, it indicates a higher accuracy.

\subsection{Model Performance for Authorized Users}
\begin{figure*}[t]

\scalebox{0.8}[0.8]{
    
    \begin{tabular*}{50mm}{@{\extracolsep{\fill}}c|c|ccc}
        Original&Ground Truth&Baseline&Correct ($K$)&Incorrect ($K'$) \\
        \begin{minipage}{4truecm}
             \centering
              \includegraphics[bb=0 0 2048 1024,scale=0.055]{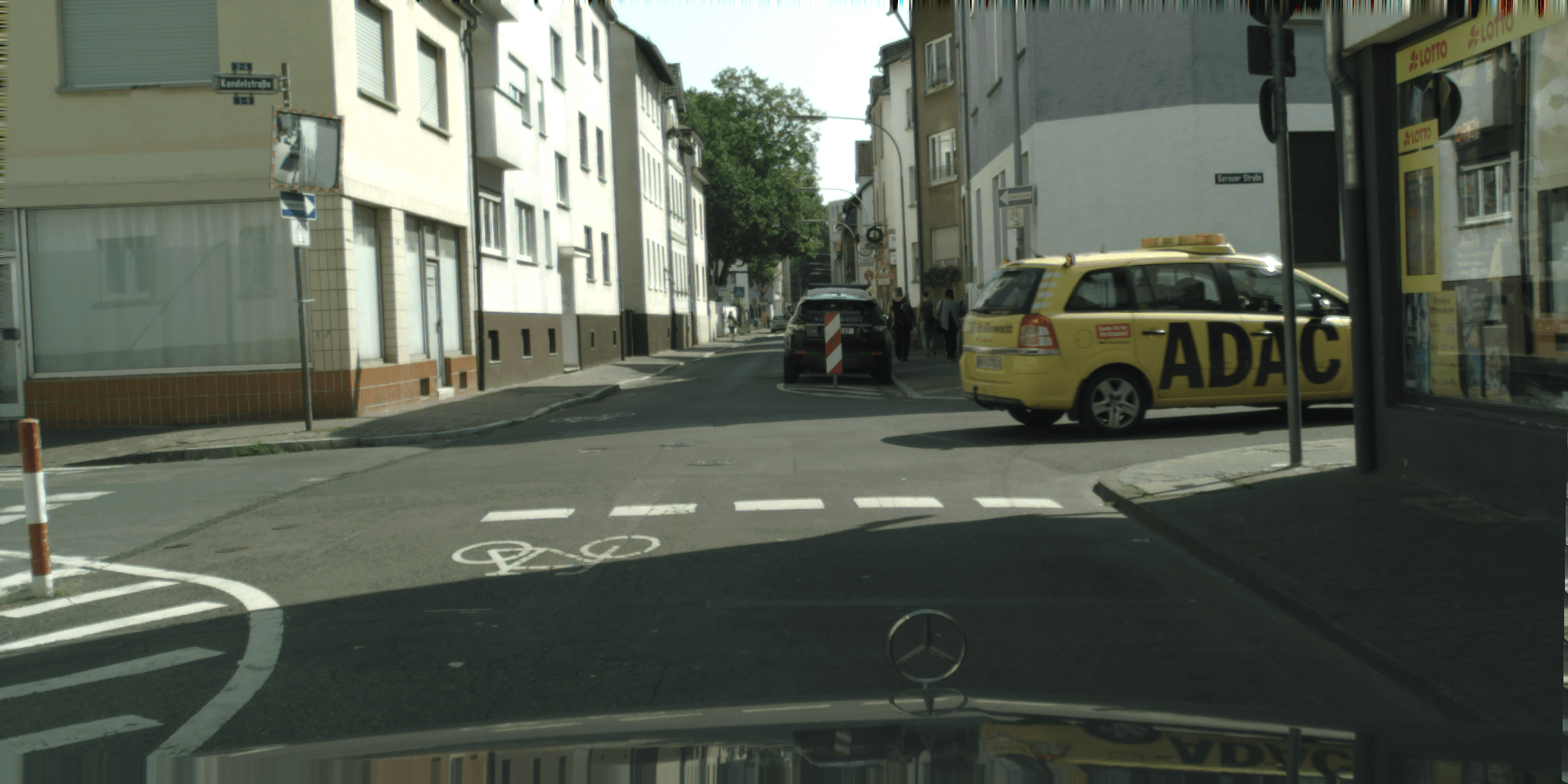}
            \end{minipage}
        &
        \begin{minipage}{4truecm}
             \centering
              \includegraphics[bb=0 0 2048 1024,scale=0.055]{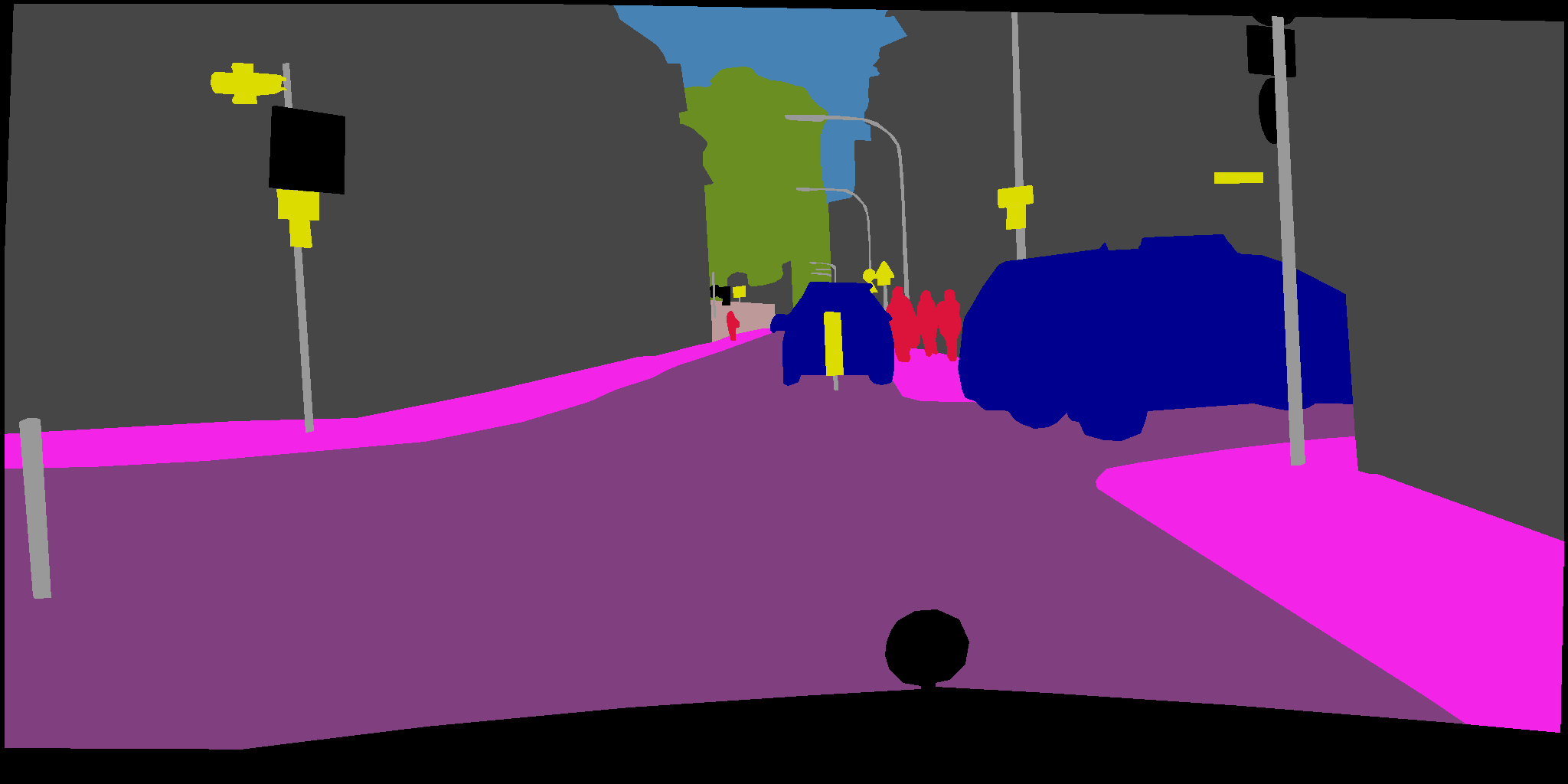}
            \end{minipage}
        &
        \begin{minipage}{4truecm}
             \centering
              \includegraphics[bb=0 0 2048 1024,scale=0.055]{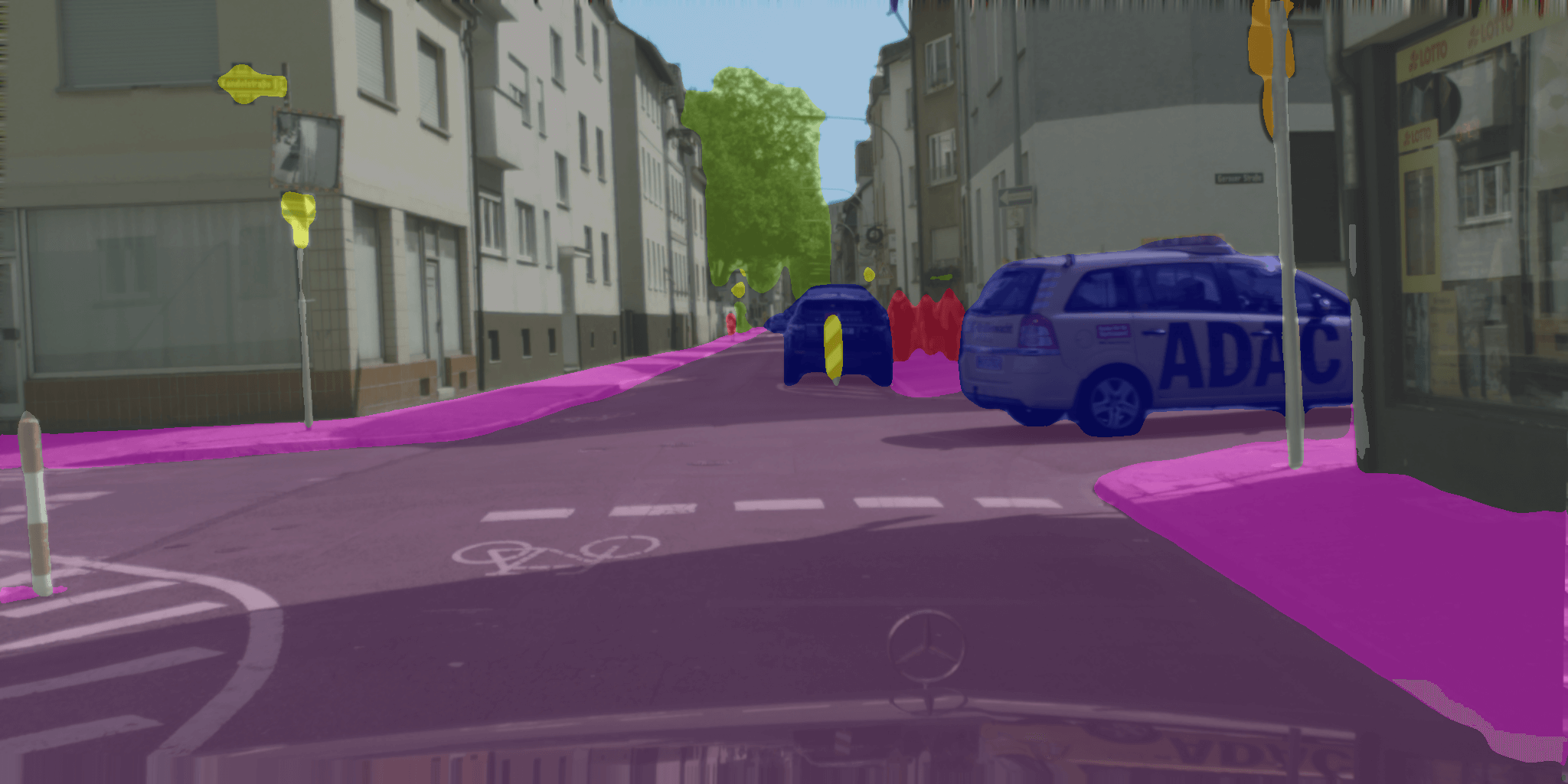}
            \end{minipage}
        &
        \begin{minipage}{4truecm}
             \centering
              \includegraphics[bb=0 0 2048 1024,scale=0.055]{Figure/1/ok_frankfurt_000000_000294_leftImg8bit.png}
            \end{minipage}
        &
        \begin{minipage}{4truecm}
            \centering
              \includegraphics[bb=0 0 2048 1024,scale=0.055]{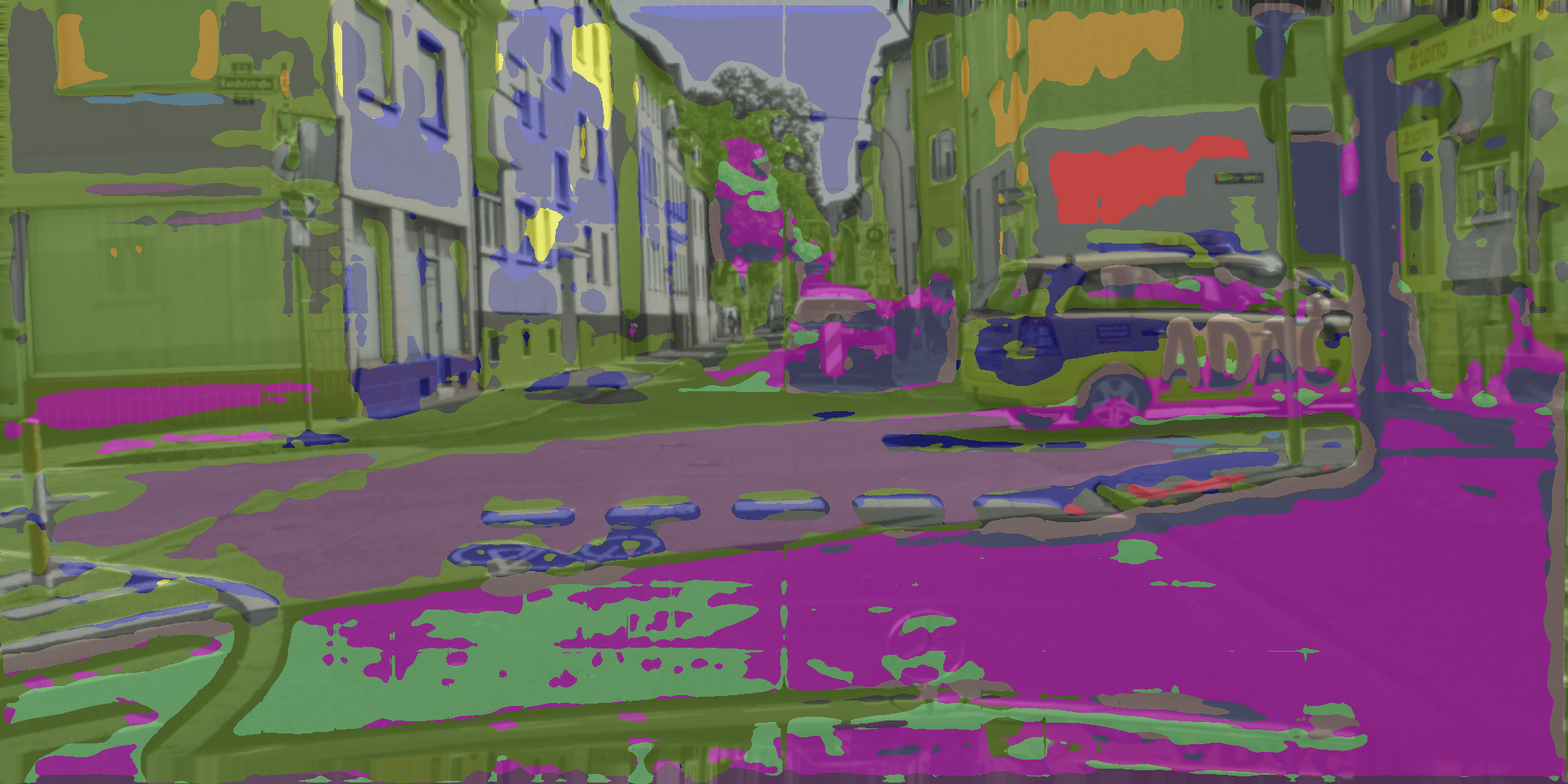}
            \end{minipage}\\
        \begin{minipage}{4truecm}
             \centering
              \includegraphics[bb=0 0 2048 1024,scale=0.055]{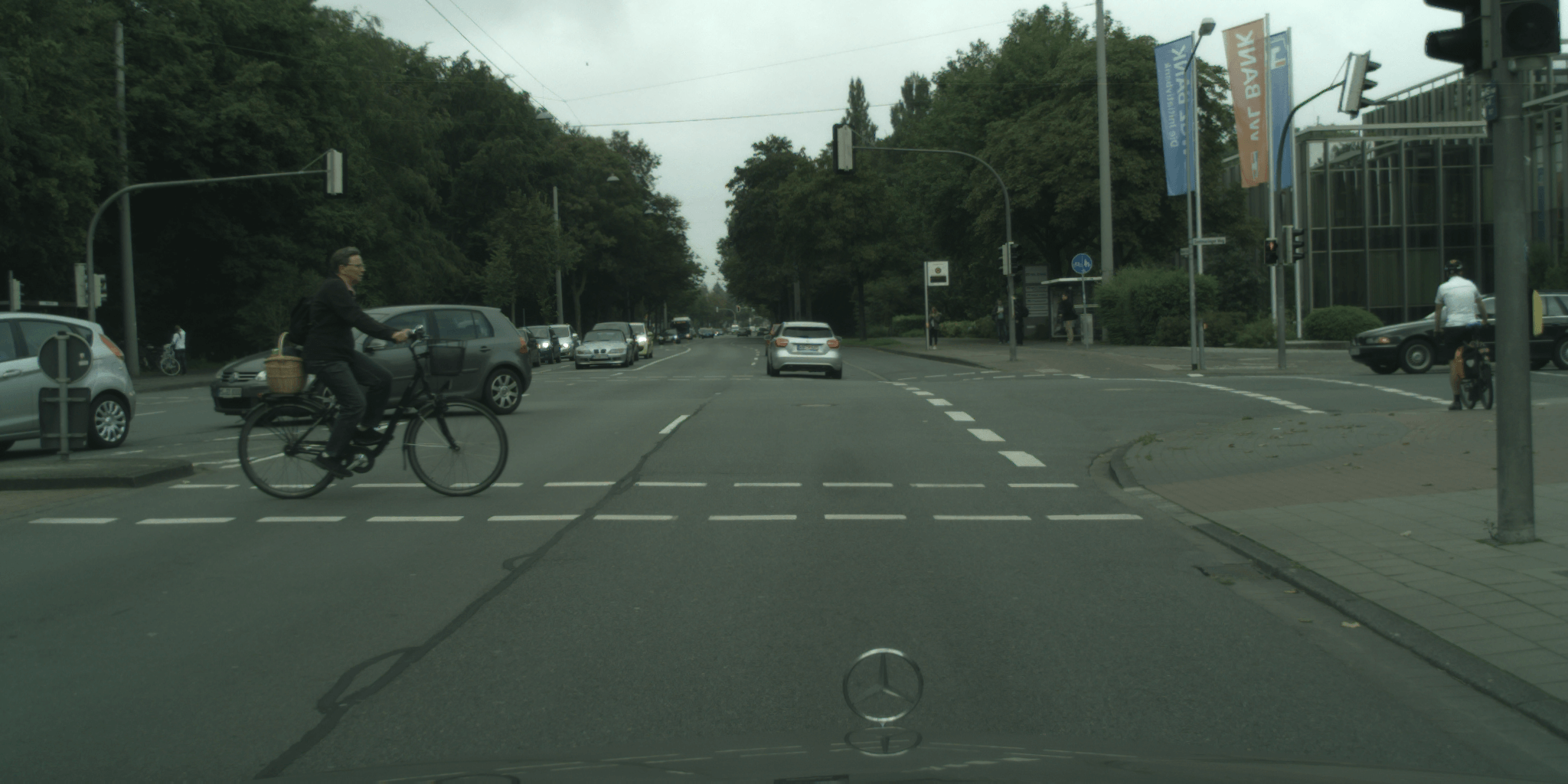}
            \end{minipage}
        &
        \begin{minipage}{4truecm}
             \centering
             \includegraphics[bb=0 0 2048 1024,scale=0.055]{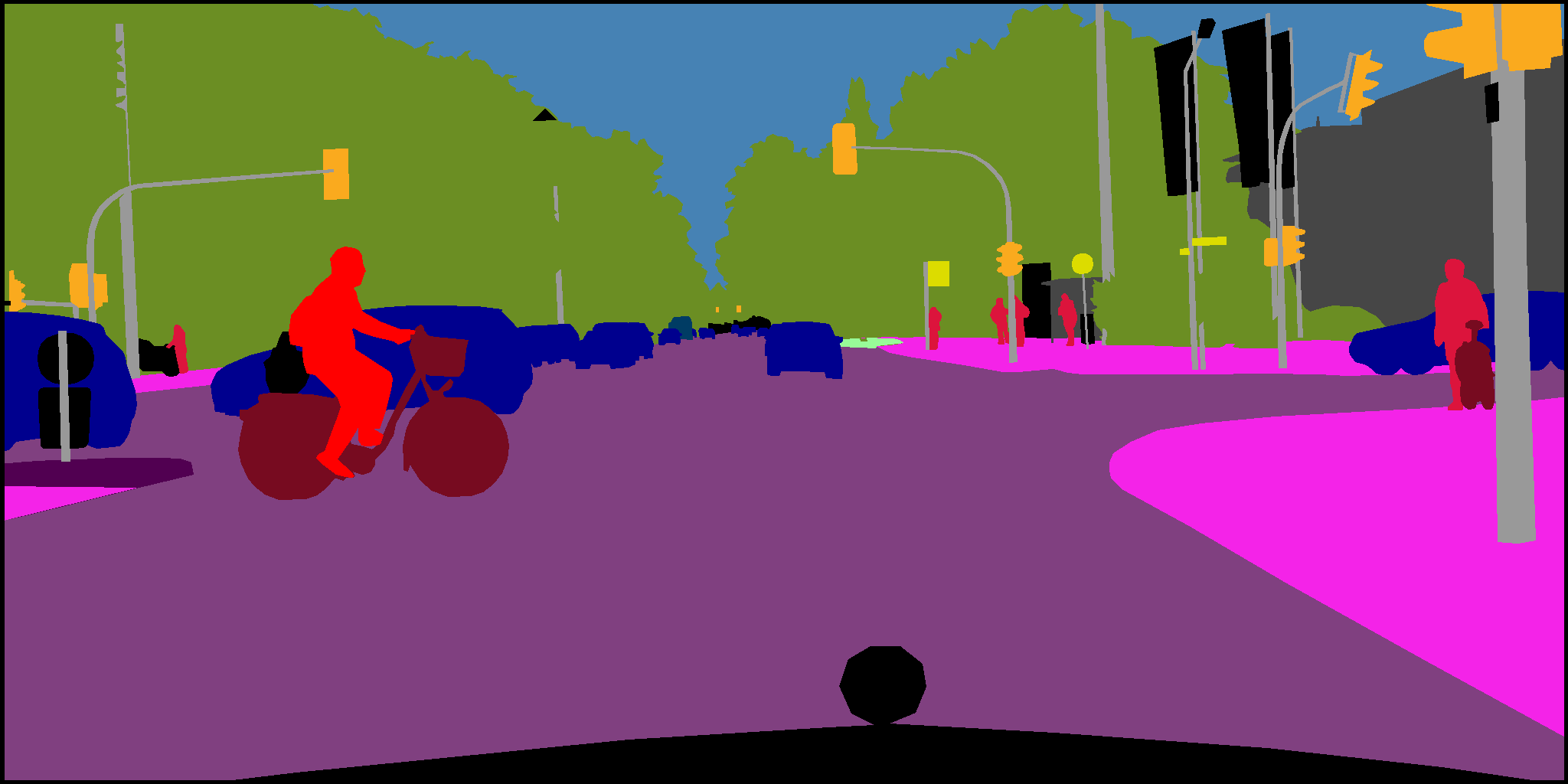}
            \end{minipage}
        &
        \begin{minipage}{4truecm}
             \centering
              \includegraphics[bb=0 0 2048 1024,scale=0.055]{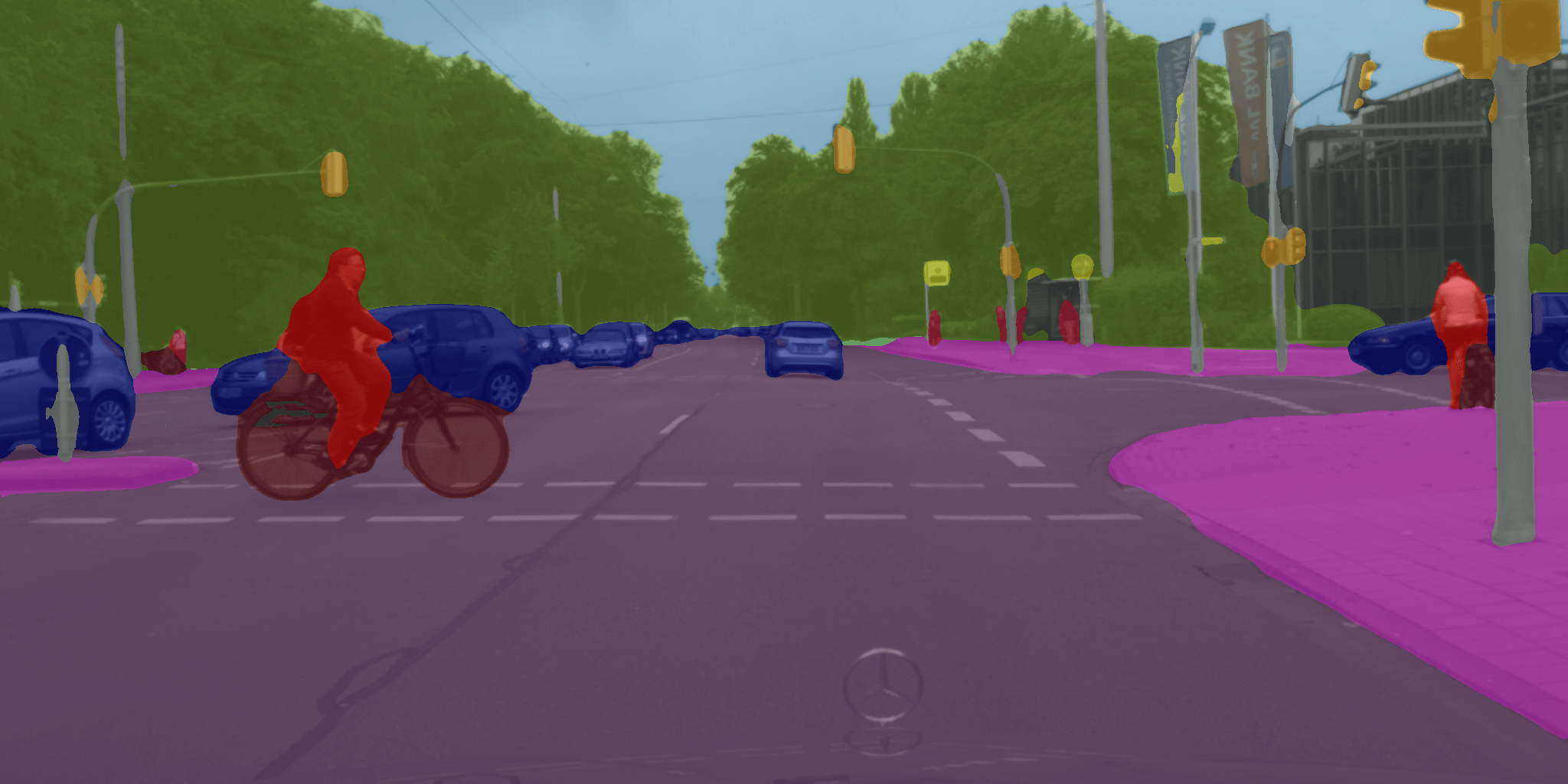}
            \end{minipage}
        &
        \begin{minipage}{4truecm}
             \centering
              \includegraphics[bb=0 0 2048 1024,scale=0.055]{Figure/2/ok_munster_000000_000019_leftImg8bit.png}
            \end{minipage}
        &
        \begin{minipage}{4truecm}
             \centering
              \includegraphics[bb=0 0 2048 1024,scale=0.055]{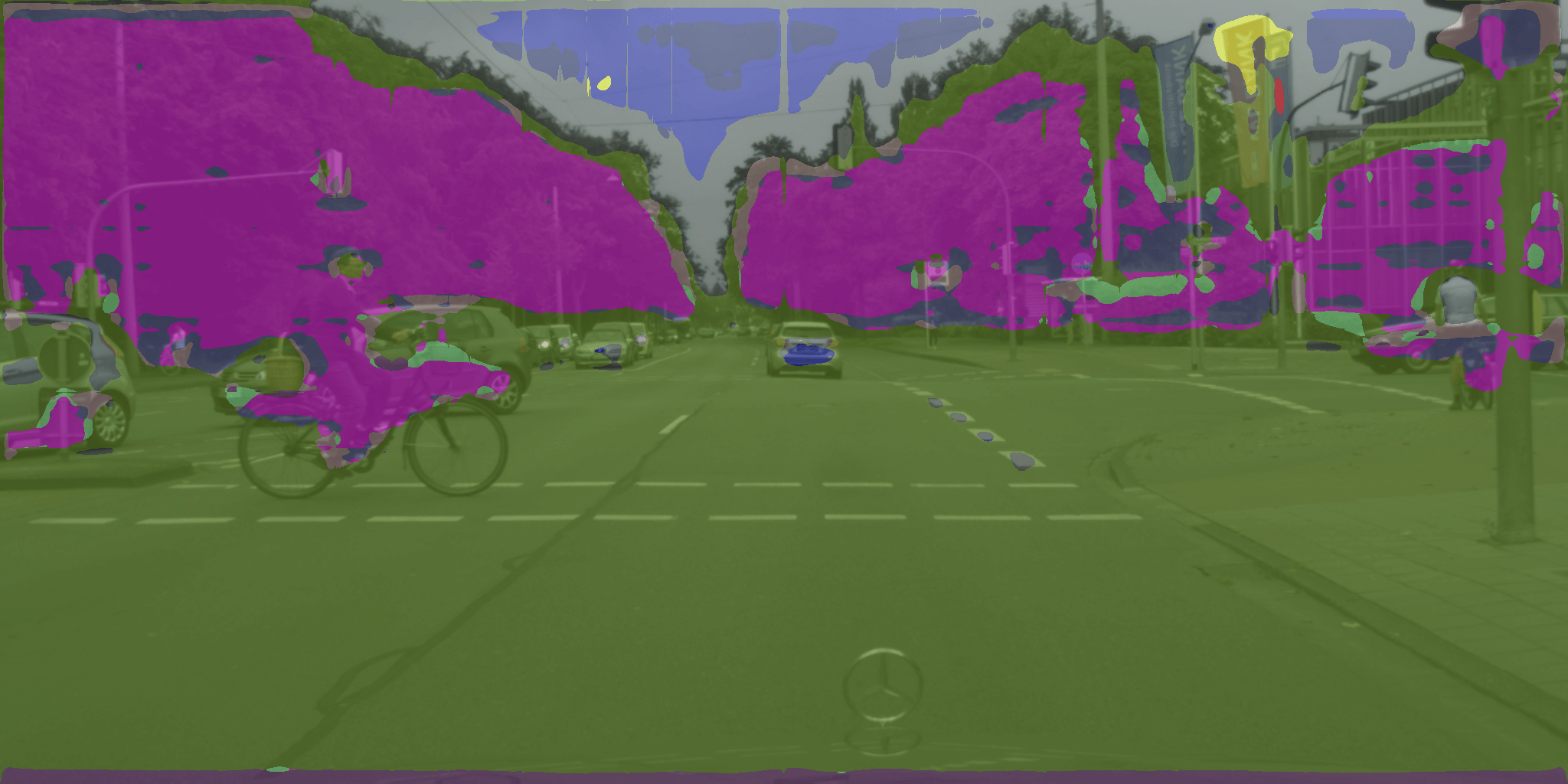}
            \end{minipage}\\
            
        \begin{minipage}{4truecm}
             \centering
              \includegraphics[bb=0 0 2048 1024,scale=0.055]{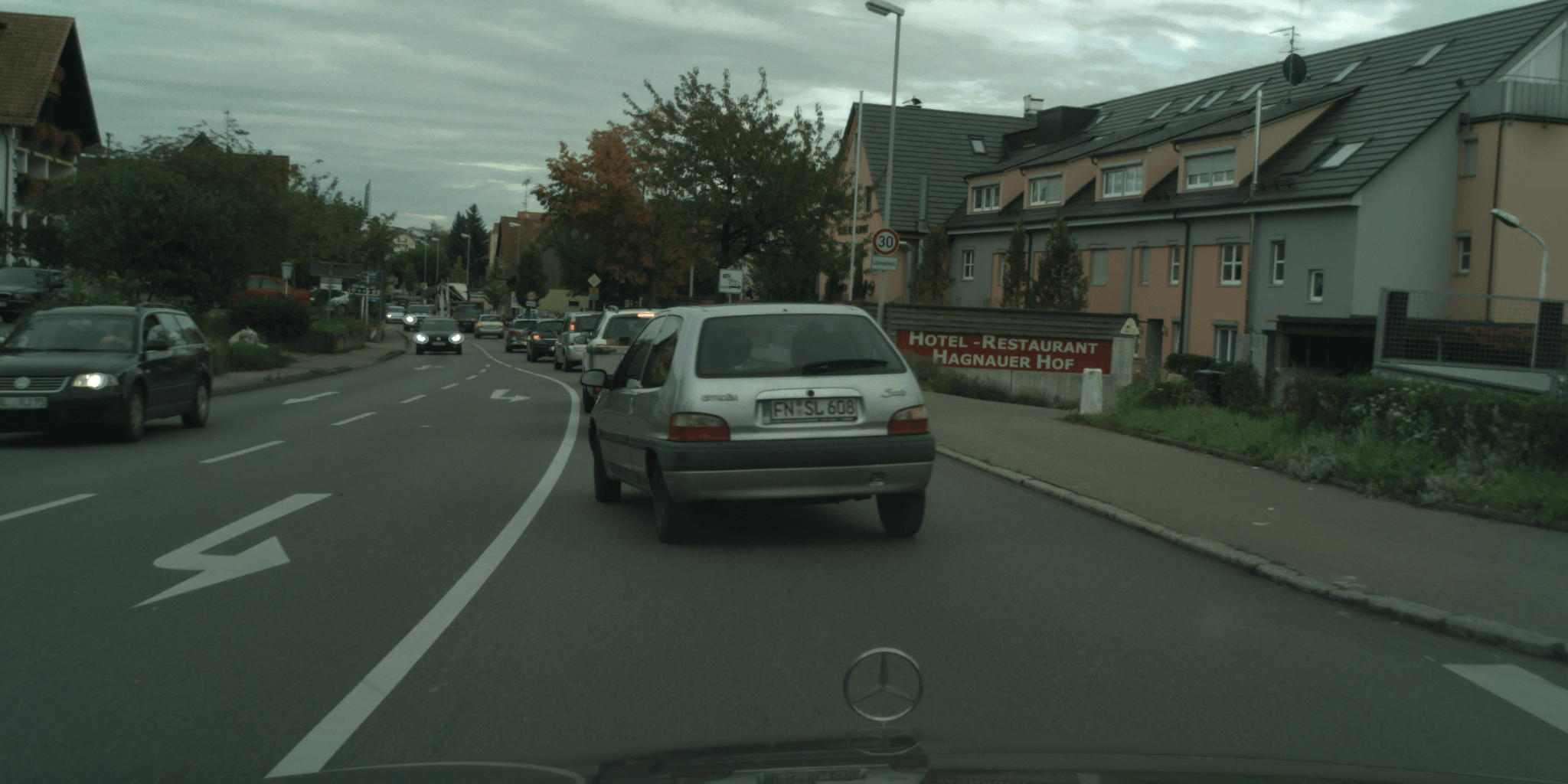}
            \end{minipage}
        &
        \begin{minipage}{4truecm}
             \centering
             \includegraphics[bb=0 0 2048 1024,scale=0.055]{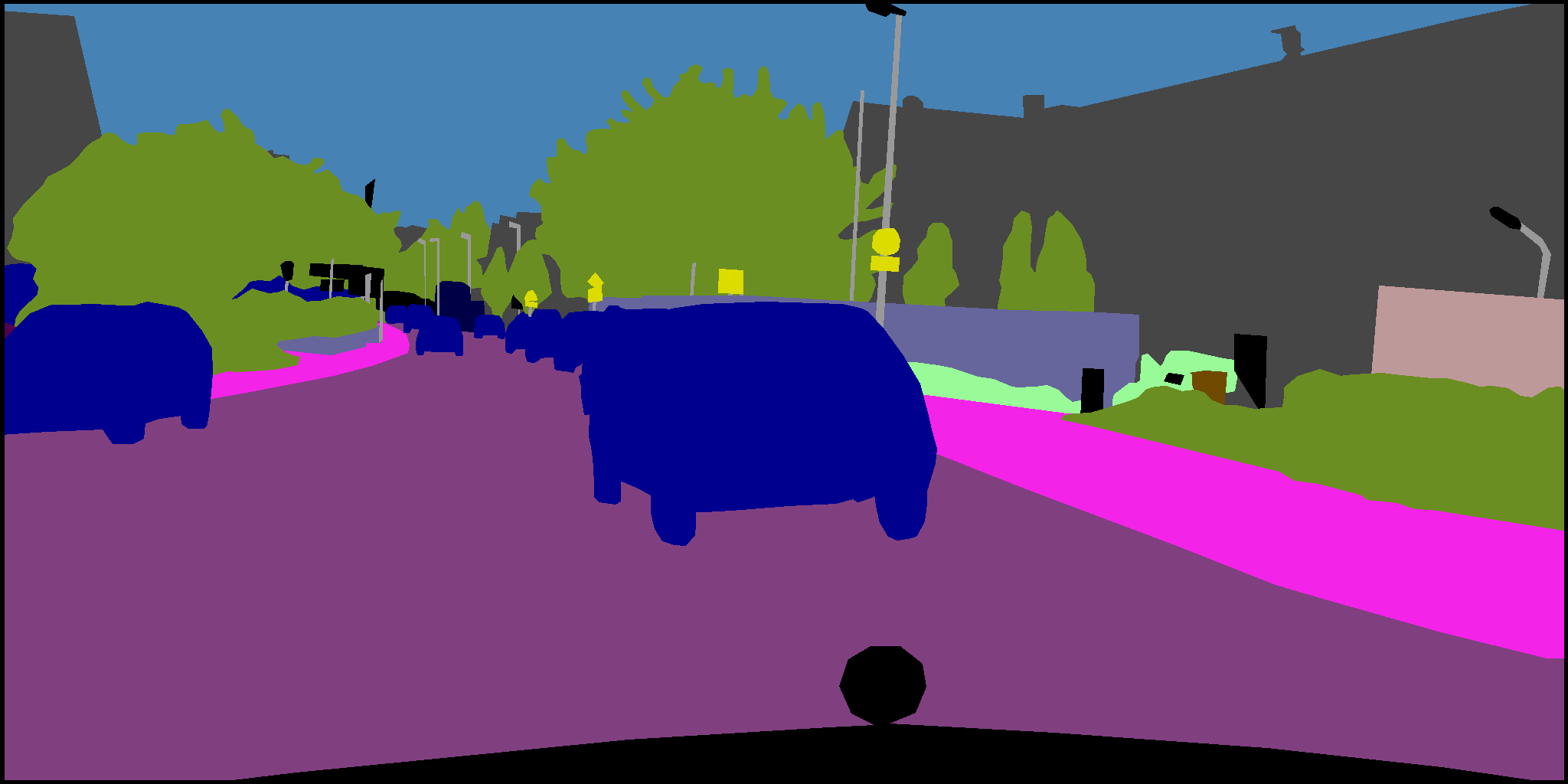}
            \end{minipage}
        &
        \begin{minipage}{4truecm}
             \centering
              \includegraphics[bb=0 0 2048 1024,scale=0.055]{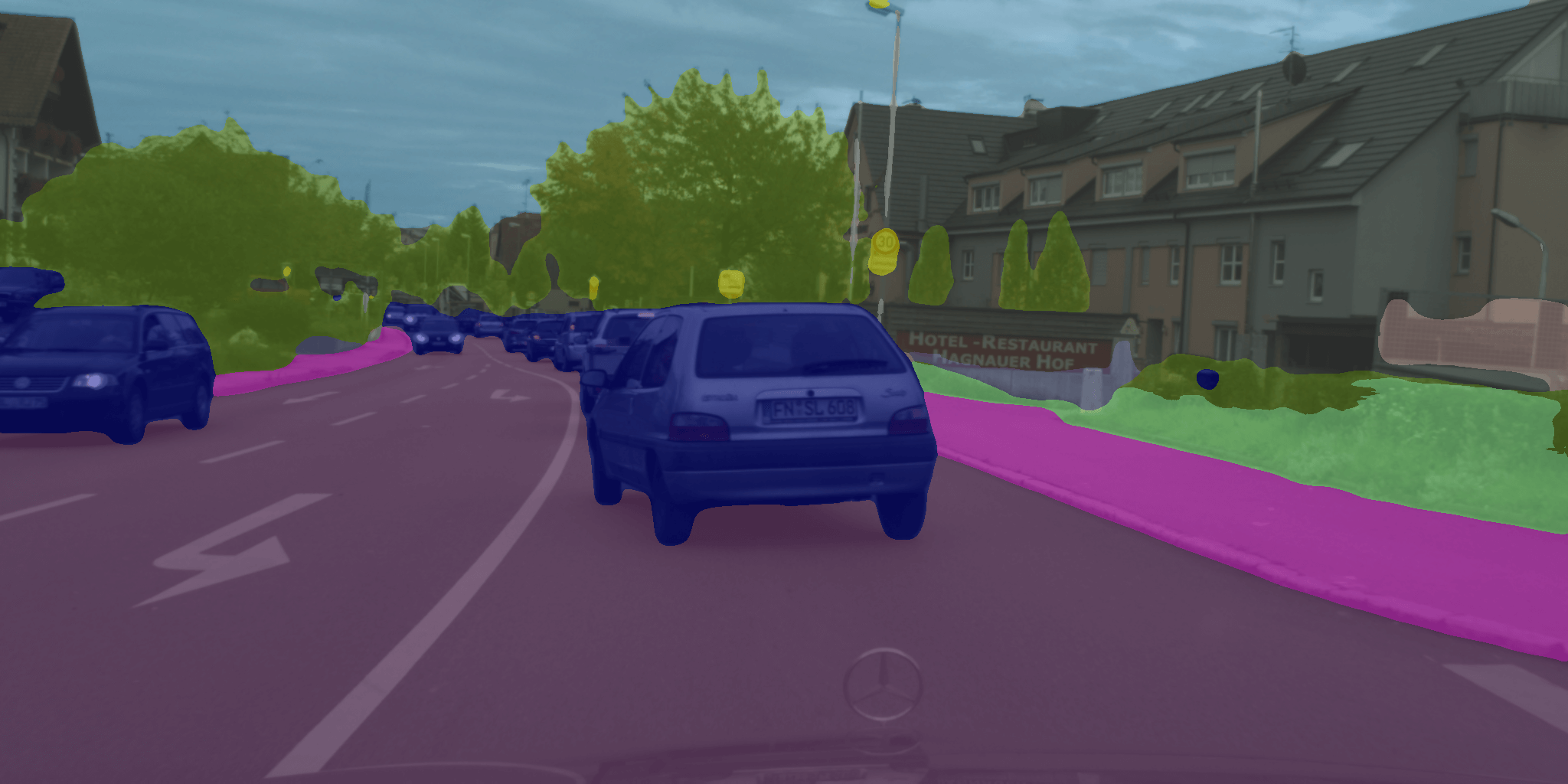}
            \end{minipage}
        &
        \begin{minipage}{4truecm}
             \centering
              \includegraphics[bb=0 0 2048 1024,scale=0.055]{Figure/3/ok_lindau_000000_000019_leftImg8bit.png}
            \end{minipage}
        &
        \begin{minipage}{4truecm}
             \centering
              \includegraphics[bb=0 0 2048 1024,scale=0.055]{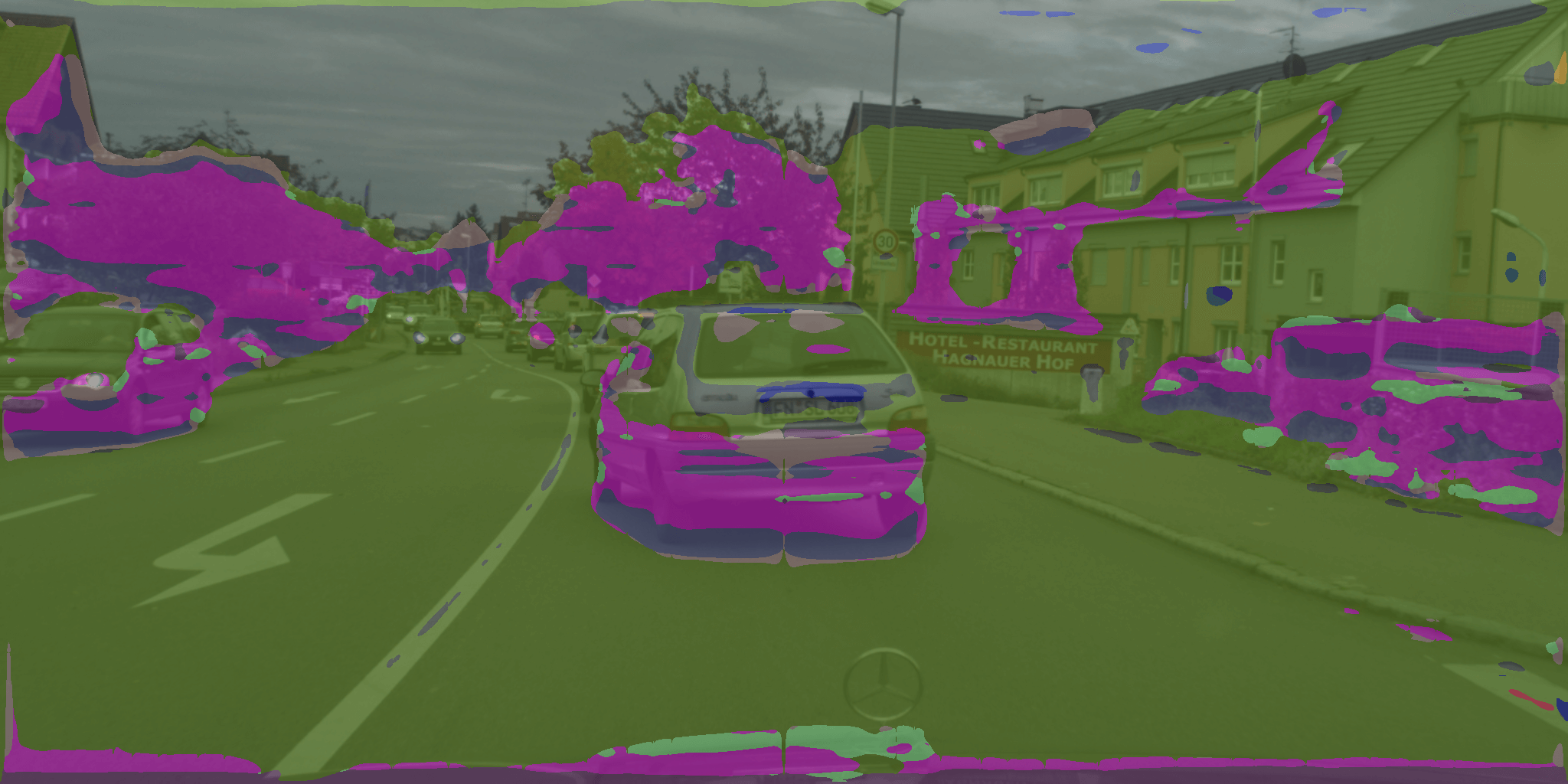}
            \end{minipage}\\

    \end{tabular*}
    }
    \caption{An example of predicted segmentation maps (Cityscapes-PUP)}
    \label{segmentation maps}
\end{figure*}

In the experiment, matrix  \textbf{E} was encrypted by using $\mathbf{E_{enc}}$ generated with a secret key $K$. Table \ref{table:result} shows experimental results for each decoder on the two datasets. Correct($K$) represents results when a user used key $K$ to encrypt test images. Baseline represents the results of models without encryption. Fig. \ref{segmentation maps} shows an example of predicted segmentation maps. Table \ref{table:result} and Fig. \ref{segmentation maps} show that there was no accuracy degradation when users had key $K$.

\begin{table}[bt]
 \caption{Accuracy of proposed models (mIoU)}
 \label{table:result}
 \centering
 \scalebox{0.78}{
  \begin{tabular}{|c|c|ccc|}
   \hline
   Dataset  & Selected decoder &Baseline&Correct ($K$) & Incorrect ($K'$) \\
   \hline
   \multirow{3}{*}{Cityscapes} & \textit{Na\"{i}ve} & 0.6457 & \textbf{0.6457} & 0.0240 \\
                               &\textit{MLA} & 0.7195 & \textbf{0.7195} & 0.0119 \\
                               &\textit{PUP}& 0.7550 & \textbf{0.7550} & 0.0201 \\
   \hline
   \multirow{3}{*}{ADE20K} & \textit{Na\"{i}ve} & 0.3527 & \textbf{0.3527} & 0.0027 \\
                           & \textit{MLA} & 0.4198 & \textbf{0.4198} & 0.0021 \\
                           & \textit{PUP}& 0.4153 & \textbf{0.4153} & 0.0022 \\
   \hline
  \end{tabular}
  }
\end{table}

\begin{figure*}[tb]
    \centering
        \begin{minipage}{0.35\linewidth}
             \centering
              \includegraphics[bb=0 0 432 288,scale=0.4]{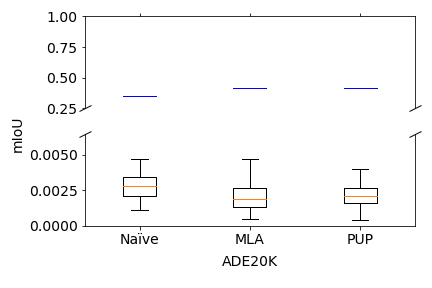}
        \end{minipage}
        \begin{minipage}{0.35\linewidth}
        \centering
        \includegraphics[bb=0 0 432 288,scale=0.4]{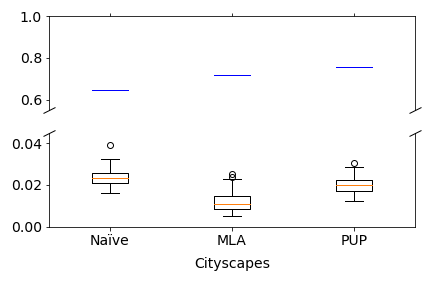}
        \end{minipage}\\
    \caption{Mean IoU ($mIoU$) values of protected models with 50 incorrect keys. Boxes span from first to third quartile, referred to as $Q_{1}$ and $Q_{3}$, and whiskers show maximum and minimum values in range of [$Q_{1} - 1.5(Q_{3} - Q_{1}), Q_{3} + 1.5(Q_{3} - Q_{1})$]. Band inside box indicates median. Outliers are indicated as dots. 
    Blue lines represent each baseline.
    }
    \label{boxplot}
\end{figure*}

\subsection{Model Performance for Unauthorized Users}
Incorrect($K'$) in Table \ref{table:result} shows the results for unauthorized users without key K used for training a model. In the experiment, we randomly generated 50 keys, and the average value of 50 trials was computed as a $mIou$ value.\par
Table \ref{table:result} and Fig. \ref{segmentation maps} show that the accuracy for unauthorized users was significantly decreased. Fig. \ref{boxplot} shows performance of the encrypted models when incorrect keys were used. In cases of unauthorized access, the highest value of mIoU in the experiments was 0.04.\par
Therefore, the proposed method was robust enough against unauthorized access, and it was confirmed to be effective as a model protection method.
\section{Conclusion}
In this paper, we proposed a protection method for segmentation transformer (SETR) models for the first time, which is semantic segmentation models with the vision transformer. From the experiment, we showed that the proposed method is robust enough to unauthorized access while maintaining a high performance for authorized users.


\bibliographystyle{IEEEtran} 
\bibliography{ref}


\end{document}